\documentclass[10pt,twocolumn,letterpaper]{article}

\usepackage{cvpr}              

\usepackage{graphicx}
\usepackage{amsmath}
\usepackage{amssymb}
\usepackage{booktabs}

\usepackage{colortbl}
\usepackage{enumitem} 
\usepackage{verbatim} 
\usepackage{xcolor}
\usepackage{multirow}
\usepackage{multicol}
\usepackage{bigstrut}
\usepackage{booktabs}
\usepackage{comment}

\usepackage{algorithm}
\usepackage{algpseudocode}
\usepackage{longtable}
\makeatletter
\DeclareRobustCommand\onedot{\futurelet\@let@token\@onedot}
\def\@onedot{\ifx\@let@token.\else.\null\fi\xspace}
\usepackage[accsupp]{axessibility}  

\usepackage{makecell}

\usepackage{xcolor}

\newcommand{\ieno}{\textit{i}.\textit{e}.}

\newcommand{\etcno}{\textit{etc.}}
\newcommand{\ours}{\textit{CLUDA-ReID}}
\newcommand{\ourtask}{\textit{LUDA person ReID}}

\usepackage[pagebackref,breaklinks,colorlinks]{hyperref}

\usepackage[capitalize]{cleveref}
\crefname{section}{Sec.}{Secs.}
\Crefname{section}{Section}{Sections}
\Crefname{table}{Table}{Tables}
\crefname{table}{Tab.}{Tabs.}

\def\cvprPaperID{1508} 
\def\confName{CVPR}
\def\confYear{2022}

\begin{document}

\title{Lifelong Unsupervised Domain Adaptive Person Re-identification \\ with Coordinated Anti-forgetting and Adaptation}

\author{
Zhipeng Huang$^{1\dagger}$\thanks{This work was done when Zhipeng Huang was an intern at MSRA.}\quad  
Zhizheng Zhang$^{2\dagger}$\footnotemark[3]\quad  
Cuiling Lan$^2$\footnotemark[3]\quad 
Wenjun Zeng$^3$\quad 
Peng Chu$^2$\\ 
Quanzeng You$^2$\quad 
Jiang Wang$^2$\quad 
Zicheng Liu$^2$\quad 
Zheng-jun Zha$^1$\\
$^1$University of Science and Technology of China \\
$^2$Microsoft \quad
$^3$EIT Institute for Advanced Study \\
{\tt\small hzp1104@mail.ustc.edu.cn\quad zengw2011@hotmail.com\quad zhazj@ustc.edu.cn}\\ 
{\tt\small \{zhizzhang, culan, pengchu, quyou, jiangwang, zliu\}@microsoft.com} 
}
\maketitle
\renewcommand{\thefootnote}{\fnsymbol{footnote}}
\footnotetext[2]{Equal contribution.} 
\footnotetext[3]{Corresponding authors.}

\begin{abstract}
   Unsupervised domain adaptive person re-identification (ReID) has been extensively investigated to mitigate the adverse effects of domain gaps. Those works assume the target domain data can be accessible all at once. However, for the real-world streaming data, this hinders the timely adaptation to changing data statistics and sufficient exploitation of increasing samples. In this paper, to address more practical scenarios, we propose a new task, Lifelong Unsupervised Domain Adaptive (LUDA) person ReID. This is challenging because it requires the model to continuously adapt to unlabeled data in the target environments while alleviating catastrophic forgetting for such a fine-grained person retrieval task. We design an effective scheme for this task, dubbed CLUDA-ReID, where the anti-forgetting is harmoniously coordinated with the adaptation. Specifically, a meta-based Coordinated Data Replay strategy is proposed to replay old data and update the network with a coordinated optimization direction for both adaptation and memorization. Moreover, we propose Relational Consistency Learning for old knowledge distillation/inheritance in line with the objective of retrieval-based tasks. We set up two evaluation settings to simulate the practical application scenarios. Extensive experiments demonstrate the effectiveness of our CLUDA-ReID for both scenarios with stationary target streams and scenarios with dynamic target streams.
\end{abstract}

\section{Introduction}
\label{sec:intro}

\begin{figure}[!t]
	\begin{center}
		\includegraphics[width=1.0\linewidth]{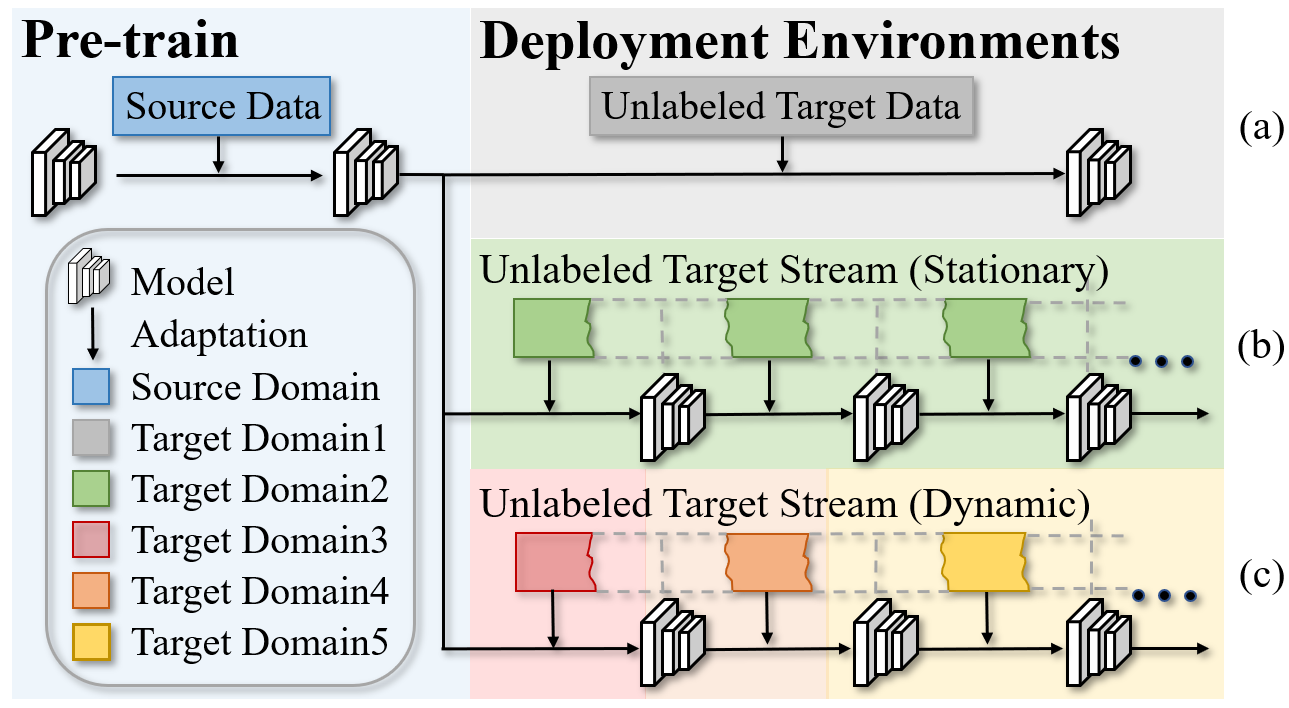}
	\end{center}
	\vspace{-0.7cm}
	\caption{Illustration of the regular UDA (a), the lifelong UDA in the stationary target scenario (b) and dynamic target scenario (c).}
	\vspace{-0.5cm}
	\label{fig:intro}
\end{figure}

Person re-identification (ReID) aims to identify the same person across different locations, time instances and cameras. It is of high value in a wide range of applications~\cite{wang2013intelligent} in diverse environments, such as customer activity analysis, missing children finding. In practice, a trained ReID model usually suffers from severe performance drops when deployed in new environments due to the domain gaps between the training data and the data in new environments. Unsupervised domain adaptive (UDA) person ReID has been widely studied to help deployed person ReID models better adapt to new environments, by transferring learned knowledge from labeled source domain to unlabeled target domain \cite{zhang2019self,fu2019self,ge2020mutual,song2020unsupervised,zhang2021unsupervised,bai2021unsupervised}.

Existing UDA person ReID methods \cite{zhang2019self,fu2019self,ge2020mutual,song2020unsupervised,zhang2021unsupervised,bai2021unsupervised} assume that the data in the target domain can be accessible all at once, which does not accord well with practical scenarios. In deployment environments, new people are often showing up constantly and the data statistics may vary over time. This imposes strong demands in developing a practical person ReID model/system that can support 1) flexible adaptation to data statistics on the fly so that we can timely obtain adapted performance without waiting for a long time to collect massive data and 2) continuous knowledge acquisition from increasing data.
Recent works begin to look into lifelong (supervised) person ReID~\cite{wu2021generalising,pu2021lifelong}. They greatly advance practical applications by supporting continuous supervised domain adaptation, but still leave much exploration room in the following aspects. First, these works require costly annotations in the target domain. Besides human labour efforts, data labelling is time consuming, which is not conductive to quick adaptation and privacy protection.
Second, they all aim to balance anti-forgetting and adaptation, lacking explicit consideration for their coordination.
In fact, anti-forgetting can facilitate flexible/quick adaptation when harnessed in the right manner.
In this paper, we propose and study a more practical task, \ieno, Lifelong Unsupervised Domain Adaptive (LUDA) person ReID, which does not require annotations in the target domain.

\ourtask~enables deployed models to achieve continuous domain adaptation using unlabeled target stream. 
To the best of our knowledge, we are the first to define this task.
Different from conventional life-long learning \cite{li2017learning,verwimp2021rehearsal,hu2021distilling}, our proposed task is challenging attributable to two reasons. First, person ReID is a fine-grained retrieval task. Critical to this task is to capture the inter-instance relations for accurately ranking instances.
Second, in the context of lifelong learning, person ReID requires higher generalization ability on unseen identities/categories since there are constantly new people showing up, beyond the requirements for remembering how to classify seen categories as in classification tasks. Thus, for \ourtask, we expect that anti-forgetting can promote adaptation by avoiding over-fitting new data and facilitating learning more generalizable features so that we can achieve fast adaptation especially when the target domain changes or reappears.

In this paper, we design an effective scheme for \ourtask~in which anti-forgetting and adaptation are coordinated, dubbed \ours. In previous studies \cite{chaudhry2018riemannian,wu2021striking,mai2021online,delange2021continual}, anti-forgetting is commonly placed into a joint learning framework together with the training for adaptation, lacking explicit considerations on their coordination. We propose Coordinated Data Replay (CDR) to solve this problem. Particularly, we store old data with a modified ID-wise reservoir sampling algorithm, and design a meta-based optimization strategy to align the objective of anti-forgetting with that of adaptation when replaying stored old data. Moreover, we maintain a model that is timely updated and a momentum-updated model that accumulates learned knowledge over time. 
We further introduce Relational Consistency Learning (RCL) to promote knowledge distillation/inheritance from the historical model to the current model when performing adaptation. Different from other knowledge distillation strategies \cite{wu2021generalising} in conventional lifelong learning, RCL is carefully designed to match the task characteristics of person ReID. We introduce two scenarios with stationary or dynamic target streams as illustrated in Fig.\ref{fig:intro}, and evaluate our \ours~ on these settings. Besides, we build a new dataset named MMP-Retrieval upon the MMPTRACK dataset for unseen-domain generalization evaluation, released at \url{https://iccv2021-mmp.github.io/subpage/dataset.html}.

Our contributions can be summarized in three aspects:

\begin{itemize}[noitemsep,nolistsep,leftmargin=*]
\item We propose a novel yet realistic task, Lifelong Unsupervised Domain Adaptive (LUDA) person re-identification, to endow person ReID models with automatic adaptation using continuously collected unlabeled data. 
\item We design an effective \ourtask~scheme, \ieno, \ours, wherein Coordinated Data Replay (CDR) and Relational Consistency Learning (RCL) are proposed to explicitly coordinate anti-forgetting and adaptation.
\item We set up two practical evaluation schemes to simulate real application scenarios. Extensive experiments demonstrate that our proposed method is effective in achieving LUDA for scenarios with stationary or dynamic target streams and enhancing unseen-domain generalization.
\end{itemize}

\section{Related Works}
\label{sec:related}

\subsection{Domain Adaptive Person Re-identification}

Person re-identification (ReID) has been widely investigated and applied in many real-world scenarios.
Unsupervised domain adaptive (UDA) person ReID is of high practical values in transferring the knowledge from labeled source domain to unlabeled target domain for improving the accuracy on the target domain after supervised fine-tuning.
Approaches in this field can be grouped into three main categories: (1) Style translation based methods \cite{liu2019adaptive,ge2020structured}; (2) Pseudo labelling based methods \cite{yu2019unsupervised,ge2020mutual,zhai2020multiple,zhao2020unsupervised}; (3) Domain-invariant feature learning based methods \cite{lin2018multi,huang2019domain,liu2020domain}. Albeit their effectiveness, they all assume that the target domain can be entirely accessible but this may not hold in practice. In the real-world, target data usually comes in a stream and its statistics may vary over time.

Recently, a few works \cite{wu2021generalising,pu2021lifelong} study the lifelong (supervised) person ReID problem wherein supervised domain adaptation is continually performed. In \cite{wu2021generalising}, consistencies from different perspectives (\ieno, classification, distribution and representation) are incorporated into a comprehensive training objective for distilling knowledge from the old model. But it does not explicitly consider the optimization in line with the objective of learning to rank. 
Pu \etal \cite{pu2021lifelong} propose an adaptive knowledge accumulation method for this task. However, both of them consider the supervised setting and lack explicit consideration on coordinating anti-forgetting with adaptation, in the sense that enabling anti-forgetting to enhance adaptation. In this work, we propose and address a new task, \ieno, \ourtask.

\subsection{Lifelong Learning}

Lifelong learning aims to learn from ever-expanding data, which is challenging in capturing new knowledge as well as maintaining old knowledge. Approaches in this field can be summarized into three categories. (1) \textit{Knowledge distillation based methods}~\cite{li2017learning,iscen2020memory,wu2021generalising} transfer already acquired knowledge from a fixed old model maintained at a certain moment to the new model currently being trained via adding consistency regularization between them. Existing works in this view ignore the design of a consistency regularization in line with the task characteristics of person ReID, \ieno, learning to rank. This problem is addressed in this paper via our proposed RCL. (2) \textit{Sample storing/generation based methods} recall old knowledge on the stored \cite{maracani2021recall,riemer2018learning,verwimp2021rehearsal} or synthetic \cite{wu2018memory,cheraghian2021synthesized} data, interleaved with the training on new collected data. They aim to strike a balance between the knowledge learned on old and new data respectively, lacking explicit consideration on the coordination between different optimization objectives at the task level. (3) \textit{Dynamic model based methods} \cite{rusu2016progressive,cortes2017adanet,xiao2014error} handle expanding data/knowledge with progressively modified model architecture. Our framework is based on combining the first two categories.

\begin{figure*}[!t]
	\centering
	\includegraphics[width=0.96\textwidth]{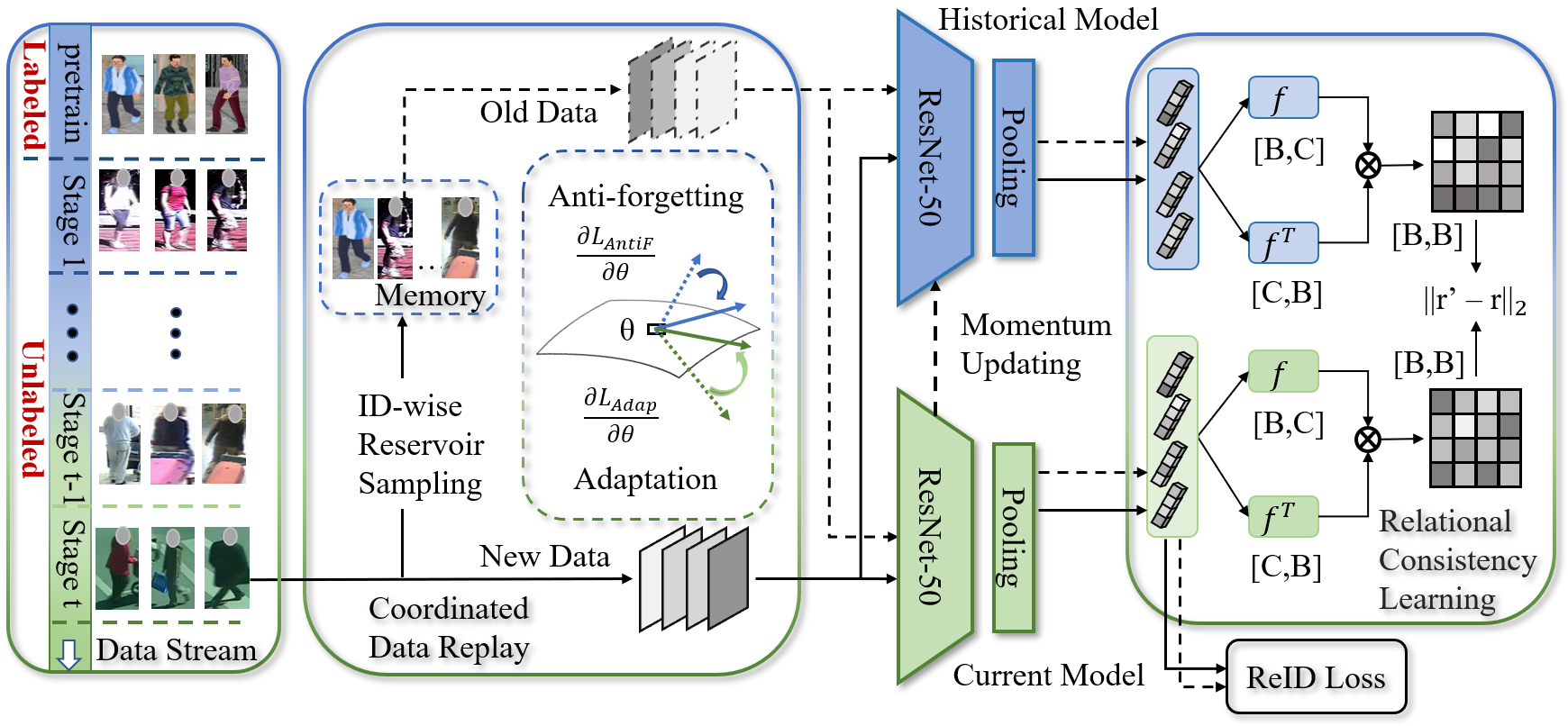}
	\caption{The overall architecture of the proposed scheme \ours~for \ourtask. In \ours, we set up a memory buffer to store samples that have been seen/trained as the old data, and preserve a momentum-updated model as the historical model to accumulate acquired knowledge over time. We fed both the old data (dashed lines) and new data (solid lines) into the current model and the historical model for capturing new knowledge for timely adaptation while enhancing the memorization of old knowledge. We achieve coordinated anti-forgetting and adaptation learning via our proposed Coordinated Data Replay and Relational Consistency Learning. }
	\label{fig:framework}
\end{figure*}

\section{Proposed Problem: LUDA Person ReID}
\label{sec:problem}

\subsection{Problem Definition}
\label{subsec:definition}

Consider a realistic scenario: a person ReID model or system is offline trained on labeled source data, then deployed in new environments wherein new data arrives continuously. Usually, the new data has domain gaps with the source data and is hard to be annotated in time. \textit{Can we utilize such unlabeled new data to perform continual domain adaptation so that the deployed model can quickly adapt to new environments?} This motivates the problem of LUDA person ReID.

For \ourtask~task, we assume that a labeled source dataset $\mathcal{X}^s\!=\!\{(\textbf{x}_i^s, y_i^s)|_{i=1}^{N_s}\}$ is available at first, which comprises $N_s$ samples $\textbf{x}_i^s\!\in\!\mathcal{X}^s$ with corresponding ID labels $y_i^s\!\in\!\mathcal{Y}_s$. The target stream is divided into $T$ splits over time, denoted by $\mathcal{X}^t\!=\!\{\textbf{x}_i^t|_{i=1}^{N^j_t}\}_{j=1}^T$. For an interval of time indexed by $j$ (called ``stage'' in the following), $N^j_t$ new unlabeled samples $\textbf{x}_i^t$ are collected in this stage and compose a set $\mathcal{X}_j^t$. After each interval/stage, $\mathcal{X}_j^t$ is used to automatically update the deployed person ReID model for continual adaptation, in the sense that we need not wait to collect the entire target dataset or train the model from scratch on the combination of newly collected and historical data many times.

\subsection{Learning Setups}
\label{subsec:setups}

In the real world, the statistics of target streams may vary over time in different degrees, corresponding to different scenarios as follows: (i) In the \textit{stationary target scenario}, such as fixed cameras in retail, supermarket, \etcno, only identities increase as new data comes while the domain/style characteristics maintain a relatively stable state. We can suppose that there are no distribution shifts between the $\mathcal{X}_j^t$ in different stages. (ii) In the \textit{dynamic target scenario}, such as cameras mounted in cars, drones, \etcno, the number of seen identities keeps increasing and their statistic characteristics also vary over time, leaving this setup being not only class-incremental but also  domain-incremental.

\section{Proposed Solution: A Coordinated Scheme}


\subsection{Overall Pipeline}
\label{sec:overall}

In our proposed scheme, we first pre-train the person ReID model on labeled source data $\mathcal{X}^s\!=\!\{(\textbf{x}_i^s, y_i^s)|_{i=1}^{N_s}\}$. Then, we deploy the pre-trained model in the target environment and update the model with target stream data $\mathcal{X}^t\!=\!\{\textbf{x}_i^t|_{i=1}^{N^j_t}\}_{j=1}^T$ for achieving lifelong UDA. For each sample, from $\mathcal{X}^s$ or $\mathcal{X}^t$, we first use a backbone network to extract its corresponding feature tensor, then adopt a global spatial GeM pooing \cite{radenovic2018fine} on this feature tensor to obtain a feature vector as its final ID representation. As common practices in this field \cite{luo2019bag,zhang2020relation,he2020fastreid}, we employ a triplet loss and a cross-entropy based classification loss (each ID is taken as one category) as person ReID losses on the extracted feature vector for training.
In the $j$-th stage, to acquire new knowledge for unsupervised domain adaptation, we train a timely updated model (called ``current model'' for brevity) on $\mathcal{X}_j^t\!=\!\{\textbf{x}_i^t|_{i=1}^{N^j_t}\}$ with person ReID losses. Here, we generate or update the pseudo ID labels for $\mathcal{X}_j^t$ via a clustering algorithm every few training epochs.

Meanwhile, we recall old knowledge via data replay and knowledge distillation for anti-forgetting. To this end, we set up a memory buffer with a limited size for data replay and maintain a historical model for knowledge distillation. Specifically, we sample historical data stored in the memory to update the current model together with new data in each iteration, and store new samples into the memory at the end of each stage. Besides, we perform knowledge distillation from the maintained historical model to the current model by adding consistency constraints on their outputs.

\subsection{Coordinated Data Replay}
\label{sec:CDR}

Recalling old knowledge (for anti-forgetting) and capturing new knowledge (for adaptation) correspond to two different objectives, thus can be viewed as two ``tasks''. However, as we discussed in our introduction, these two tasks do not have to conflict with each other. In fact, they can be coordinated to enhance the model's generalization capability for unseen identities and new domains.
Thus, we propose Coordinated Data Replay (CDR) to coordinate the optimization of anti-forgetting and adaptation. 

The trainable network parameters of a \ourtask~model comprise a backbone network $\theta$ (such as the ResNet-50 in Fig.\ref{fig:framework}) for ReID feature extraction, and a classifier $\phi$ for adding the classification loss. In the learning stage of lifelong unsupervised domain adaptation, the optimization objective of \textbf{adaptation} can be formulated as: 
\begin{equation}\label{eq:1}
    \min_{\theta, \phi}{\mathcal{L}_{Adap}(\theta, \phi)} = \min_{\theta, \phi}{\mathcal{L}_{Tri}(\theta) + \mathcal{L}_{Cls}(\theta, \phi)},
\end{equation}
where $\mathcal{L}_{Tri}(\theta)$ is the triplet loss applied to the extracted feature vectors of new data, $\mathcal{L}_{Cls}(\theta, \phi)$ denotes the Cross-Entropy based classification loss applied to the logits of new data. Here, the new data refers to the training batch sampled from $\mathcal{X}_j^t$ in the $j$-th stage. 

To recall previously learned knowledge, we set up a limited-size memory buffer for storing data to replay these samples. This memory is updated at the end of each stage with a modified reservoir sampling algorithm. The classical reservoir sampling algorithm \cite{vitter1985random} is designed to randomly keep $N$ samples in memory from a sequence, with equal probability for each one. Here, to preserve as diverse ID information as possible, we modify the reservoir algorithm from an instance-wise one to an ID-wise one. (See more details in the supplementary.) With both the stored old data and the new data together, the optimization objective of \textbf{anti-forgetting} can be formulated as:
\begin{equation}\label{eq:2}
    \min_{\theta, \phi}{\mathcal{L}_{AntiF}(\theta,\phi)} = \min_{\theta, \phi}{\mathcal{L}_{KD}(\theta,\phi)\!+\!\mathcal{L}_{Tri}(\theta)\!+\!\mathcal{L}_{Cls}(\theta,\phi)},
\end{equation}
where $\mathcal{L}_{KD}(\theta)$ represents the knowledge distillation loss for transferring already acquired knowledge from the maintained historical model to the current model, applied to the inferenced results of both old and new data. We design a novel knowledge distillation loss in line with person ReID task and leave its detailed introduction in the subsequent section. $\mathcal{L}_{Tri}(\theta)$ and $\mathcal{L}_{Cls}(\theta, \phi)$ denote the triplet loss and the classification loss respectively, similar to Eq.(\ref{eq:1}). Note that they are both applied for replaying the old data stored in the memory buffer. The general optimization objective of lifelong learning with the regular data replay is:
\begin{equation}\label{eq:3}
    \min_{\theta, \phi}\ {\mathcal{L}_{Adap}(\theta, \phi)} + {\mathcal{L}_{AntiF}(\theta, \phi)}.
\end{equation}

In our proposed CDR, we aim to align two different optimization objectives as described above via a meta-optimization strategy. We draw inspiration from Model-Agnostic Meta-learning (MAML) \cite{finn2017model} which splits the optimization into meta-train and meta-test processes and then involves a meta-update with a gradient-through-gradient mechanism such that the model is updated along an aligned direction between the meta-train and the meta-test. To coordinate \textbf{adaptation} and \textbf{anti-forgetting}, we propose to treat these two tasks as meta-train and meta-test in each iteration of parameter updating rather than splitting samples into meta-train and meta-test as in \cite{li2018learning,riemer2018learning}. With the task of adaptation as meta-train while the task of anti-forgetting as meta-test, the overall meta-optimization objective is: 
\begin{equation}\label{eq:4}
\begin{aligned}
    &\min_{\theta, \phi}{\mathcal{L}_{Adap}(\theta, \phi)} + \mathcal{L}_{AntiF}(\theta - \Delta\theta, \phi - \Delta\phi), \\
    &\Delta\theta = \alpha\nabla_{\theta}\mathcal{L}_{Adap}(\theta, \phi),\ 
    \Delta\phi = \alpha\nabla_{\phi}\mathcal{L}_{Adap}(\theta, \phi),
\end{aligned}
\end{equation}
where $\alpha$ denotes the meta-learning rate. With the meta-optimization objective, the optimization of anti-forgetting is considered through the gradient calculated upon the pre-updated parameters after the meta-train process via one gradient descent step: $\theta' \gets \theta - \alpha\nabla_{\theta}\mathcal{L}_{Adap}(\theta, \phi)$ and $\phi' \gets \phi - \alpha\nabla_{\phi}\mathcal{L}_{Adap}(\theta, \phi)$. As common practices in \cite{finn2017model,dou2019domain} for the efficient implementation of meta-learning, we omit higher-order items during performing gradient back-propagation.

One natural question may raise: \textit{Why can the meta-optimization objective presented in Eq.(\ref{eq:4}) coordinate anti-forgetting  and adaptation for lifelong learning?} Inspired by the analysis in \cite{li2018learning}, we recall the first-order Taylor expansion for a general function $f(x,y)$ of two variables at $x=x_0, y=y_0$ as:
\begin{equation}\label{eq:5}
    f(x,y)\!=\!f(x_0, y_0)\!+\!f'_x(x_0, y_0)(x-x_0)\!+\!f'_y(x_0, y_0)(y-y_0),
\end{equation}
where $x_0$ is any point close to $x$. We instantiate the function $f$ as $\mathcal{L}_{AntiF}$, and set $x$ and $y$ to be $\theta-\alpha\nabla_{\theta}\mathcal{L}_{Adap}(\theta, \phi)$ and $\phi-\alpha\nabla_{\phi}\mathcal{L}_{Adap}(\theta, \phi)$, respectively. Further, we choose $x_0$ and $y_0$ to be $\theta$ and $\phi$, respectively. Thus, we can replace the second term $\mathcal{L}_{AntiF}(\theta - \Delta\theta, \phi - \Delta\phi)$ in Eq.(\ref{eq:4}) with its corresponding first-order Taylor expansion, and reformulate the overall meta-optimization objective of Eq.(\ref{eq:4}), as:
\begin{equation}\label{eq:6}
\begin{aligned}
    \min_{\theta, \phi}\ &\ {\mathcal{L}_{Adap}(\theta, \phi)} + \mathcal{L}_{AntiF}(\theta, \phi) \\
    &- \alpha\nabla_{\theta}\mathcal{L}_{Adap}(\theta, \phi)\nabla_{\theta}\mathcal{L}_{AntiF}(\theta, \phi) \\
    &- \alpha\nabla_{\phi}\mathcal{L}_{Adap}(\theta, \phi)\nabla_{\phi}\mathcal{L}_{AntiF}(\theta, \phi).
\end{aligned}
\end{equation}

Compared to the general optimization objective in Eq.(\ref{eq:3}), the above meta-optimization objective, \ieno, Eq.(\ref{eq:6}), additionally maximizes the dot product of gradients of $\mathcal{L}_{Adap}(\theta, \phi)$ and $\mathcal{L}_{AntiF}(\theta, \phi)$ with respect to $\theta$ and $\phi$. Therefore it encourages both the feature extractor and the classifier to be optimized along a consistent/aligned direction between anti-forgetting  and adaptation so that these two tasks are coordinated for more effective life-long learning.

\subsection{Relational Consistency Learning}
\label{sec:RCL}

As aforementioned, a classical method to alleviate the catastrophic forgetting problem is to transfer the knowledge from a model preserved at a certain historical time instance to the current model via knowledge distillation \cite{li2017learning,iscen2020memory,wu2021generalising}. These methods are indeed effective but leave two problems for \ourtask: 1) The task characteristics and objectives of person ReID, \ieno, learning to rank, have not been sufficiently considered in designing the loss functions for knowledge distillation; 2) The old model preserved at a certain historical moment lacks sufficient robustness for the potential training perturbation over time. We propose Relational Consistency Learning (RCL) to address these problem towards more effective knowledge distillation in line with the task characteristics of \ourtask.

We first introduce the idea of \textit{temporal ensembling} from semi-supervised learning \cite{tarvainen2017mean} into \ourtask~to build the historical model. Specifically, we initialize the historical model with the weights of source pre-trained model, then update it via a weighted summation where a momentum scalar $\beta$ is applied to historical model while $1-\beta$ is applied to the current model. 

The goal of person ReID is learning to rank. In line with this goal, we design a regular term to encourage the historical model and the current model to output consistent ranking results for input samples. In each meta-optimization step, one batch of historically stored samples (abbreviated as ``old batch'', denoted by $\mathcal{B}^{o}$) is sampled from the memory while another batch of newly collected samples (abbreviated as ``new batch'', denoted by $\mathcal{B}^{n}$) is sampled from the current stream. Both $\mathcal{B}^{o}$ and $\mathcal{B}^{n}$ are fed into the historical model and the current model to get their corresponding ID feature vectors inferred by these two models.

\begin{figure*}[!t]
	\centering
	\includegraphics[width=0.99\textwidth]{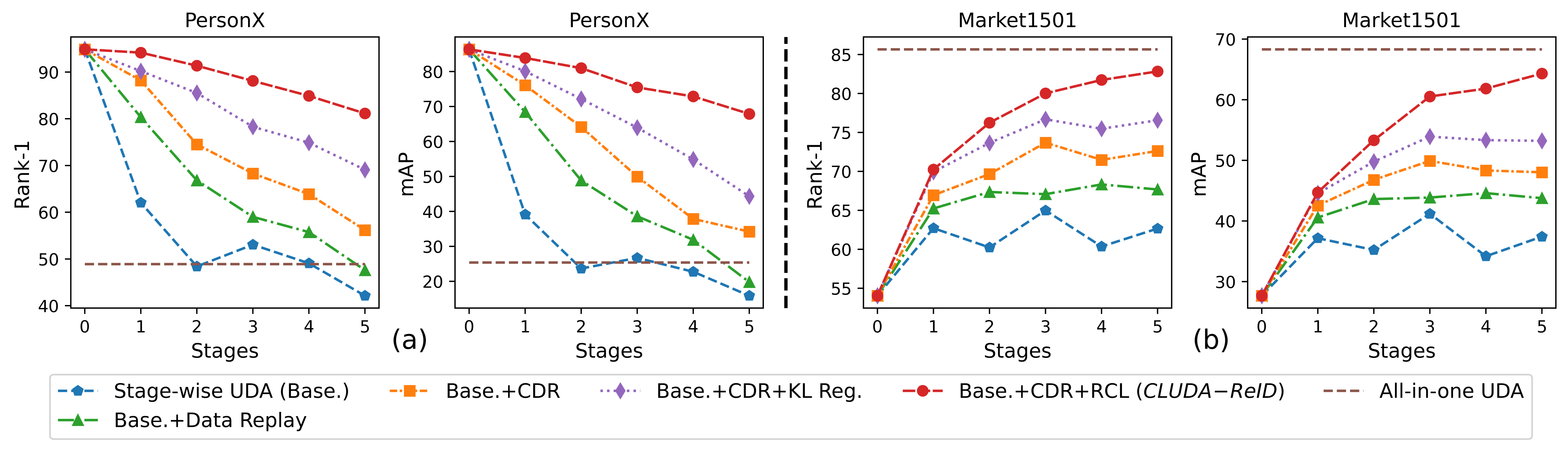}
	\vspace{-0.30cm}
	\caption{Effectiveness analysis of \ours~on the source domain (PersonX) and on the target domain (Market1501) in the stationary target scenario. In ``\textit{Stage-wise UDA (Base.)}'', we fine-tune the pre-trained model using the data collected in the current stage sequentially, without explicit considerations on anti-forgetting. In ``\textit{Base.+Data Replay}'', the data is stored into a memory buffer for replaying, wherein $\mathcal{L}_{Tri}$ and $\mathcal{L}_{Cls}$ are applied to both old data and new data for joint training, without $\mathcal{L}_{KD}$ used. ``\textit{Base.+CDR}'' denotes the model using our proposed CDR as in Eq.(\ref{eq:4}) (wherein $\mathcal{L}_{KD}$ is not adopted). In ``\textit{Base.+CDR+KL Reg.}'', we apply $\mathcal{L}_{KL}$ to both old and new data to implement $\mathcal{L}_{AntiF}$ in Eq.(\ref{eq:4}). In ``\textit{Base.+CDR+RCL}'', we further apply the proposed $\mathcal{L}_{Rel.}$ to both old and new data to achieve our proposed RCL. ``All-in-one UDA'' refers to the model combining all available target domain data for regular UDA training.}
    \vspace{-0.30cm}
	\label{fig:figure1}
\end{figure*}

We denote the feature vectors of the $i$-th and the $j$-th samples inferred by the current model as $\mathbf{f}_i^c$ and $\mathbf{f}_j^c$, respectively. Likewise, $\mathbf{f}_i^h$ and $\mathbf{f}_j^h$ represent their corresponding feature vectors extracted by the historical model. The relation/affinity between the $i$-th and the $j$-th samples inferred by the current model can be calculated by $r_{i,j}^c=\mathbf{f}_i^{c\top}\mathbf{f}_j^{c}/(\left\|\mathbf{f}_i^{c}\right\|_2\left\|\mathbf{f}_j^{c}\right\|_2)$. The relation/affinity inferred by the historical model $r_{i,j}^h$ is computed in the same way. As the core of our RCL, the regular term on inter-instance relations can be formulated as:
\begin{equation}\label{eq:7}
\mathcal{L}_{Rel.} = \sum_{i, j \in \mathcal{B}^{n}}(r_{i,j}^c-r_{i,j}^h)^2\!+\!\sum_{i', j' \in \mathcal{B}^{o}}(r_{i',j'}^c-r_{i',j'}^h)^2.
\end{equation}
We empirically find that merging the old batch $\mathcal{B}^{o}$ and the new batch $\mathcal{B}^{n}$ into one to calculate such regular term delivers very close performance to the version in Eq.(\ref{eq:7}). To be computation-efficient, we propose to calculate the affinity matrix for $\mathcal{B}^n$ and $\mathcal{B}^o$ individually. This regular term plays the regularization role in transferring the knowledge from the historical model to the current model on learning relative relations, which is required by ranking. Complementary to this, we additionally add another regular term to transfer the knowledge on learning absolute ID-related features, which is formulated as:
\begin{equation}\label{eq:8}
\mathcal{L}_{KL} = \sum_{i \in \mathcal{B}^{n}\cap\mathcal{B}^{o}}{KL}(\textbf{g}_{i}^c \| \textbf{g}_{i}^h),
\end{equation}
where $KL(\cdot\|\cdot)$ denotes the Kullback Leibler (KL) divergence distance between two distributions. $\textbf{g}_i^c\!=\!\sigma(\phi_c(\mathbf{f}_i^c))$ and $\textbf{g}_i^h\!=\!\sigma(\phi_h(\mathbf{f}_i^h))$ denotes the classification logits inferred by the current model and the historical model, respectively. Here, $\phi_c(\cdot)$ and $\phi_h(\cdot)$ are the corresponding classifiers while $\sigma(\cdot)$ is the Softmax function. As a result, in our proposed RCL, we implement the knowledge distillation loss $\mathcal{L}_{KD}(\theta,\phi)$ in Eq.(\ref{eq:2}) as: $\mathcal{L}_{KD} = \mathcal{L}_{Rel.} + \mathcal{L}_{KL}$.

\section{Experiments}
\label{sec:experiment}

\subsection{Datasets and Evaluation Metrics}
 
We employ a synthetic dataset PersonX (PX)~\cite{sun2019dissecting} for source pre-training and three public real-world image datasets Market1501 (MA)~\cite{zheng2015scalable}, CUHK-SYSU (SY)~\cite{xiao2017joint}, MSMT17 (MS)~\cite{wei2018person} for unsupervised fine-tuning. This shows a privacy-friendly practice since no ID annotations of real persons are required. Besides, for unseen-domain generalization evaluation, we build a new dataset called MMP-Retrieval upon the training and validation splits of MMPTRACK dataset released in ICCV 2021 multi-camera multiple people tracking workshop. MMP-Retrieval comprises 21 people in 5 simulated environments. 
We have released it as aforementioned.
More details of the used public datasets can be found in the supplementary. 
We use the cumulative matching characteristics (CMC) at Rank-1 (R-1) and mean average precision (mAP) for evaluation.

\subsection{Implementation Details}
Following common practices \cite{zhang2019densely,luo2019bag,he2020fastreid} in person ReID, we adopt ResNet50~\cite{he2016deep} pretrained on Imagenet \cite{deng2009imagenet} as our backbone. Similar to~\cite{luo2019bag}, the last spatial down-sampling in the ``conv5\_x'' block is removed. We resize images to 256 $\times$ 128, and set the batch size of both $\mathcal{B}^o$ and $\mathcal{B}^n$ to 64, including 16 identities and 4 images per identity for each batch. We adopt the commonly used data augmentation strategies of horizontal flipping, random cropping. Following~\cite{ge2020selfpaced,zheng2021exploiting}, we use the clustering algorithm of DBSCAN to generate pseudo labels. Unless otherwise specified, we train the model on the source domain for 60 epochs, and train the model on the stationary/dynamic target streams for 40/60 epochs per stage respectively. We adopt Adam~\cite{kingma2014adam} optimizer with its learning rate initialized as $3.5\!\times\!10^{-4}$. The size of memory buffer is set to 512. We provide the ablation study for the memory buffer size and more implementation details in the supplementary.

\subsection{Evaluation in the Stationary Target Scenario}

\noindent\textbf{Experiment configuration.}
We evaluate our proposed \ours~in the first scenario introduced in Sec.\ref{subsec:setups}. We employ the PersonX~\cite{sun2019dissecting} as the source for supervised pre-training while employing the Market1501~\cite{zheng2015scalable} for performing LUDA. We sample 750 identities of the training data of Market1501, and uniformly split them into 5 subsets for a 5-stage LUDA training. At the end of each stage, we compute the R-1 and mAP accuracy on the entire test set of the source data and the target data to respectively assess the model capacities on anti-forgetting and adaptation.

\noindent\textbf{Effectiveness of \ours.} We start with a basic model ``\textit{Stage-wise UDA}'' without anti-forgetting as the base model, then incrementally add different components upon it to assess their effectiveness. 
As shown in Fig.\ref{fig:figure1}, the performances on the source domain (PersonX) and the target domain (Market1501) reflect the \textit{anti-forgetting} ability and the timely \textit{adaptation} ability of models, respectively. Except for \textit{All-in-one UDA}, all other models are trained stage by stage as new data arrives over time. \textit{Stage-wise UDA} does not explicitly consider the optimization for anti-forgetting. It suffers from a dramatic performance drop on the source domain data and is inferior to adapting to the target domain. As the results of \textit{Base.+Data Replay} show, data replaying boosts the performance on both source and target data by revisiting old data. Relative to \textit{Stage-wise UDA}, the model \textit{Base.+CDR} achieves significant improvements. This demonstrates our proposed CDR effectively promotes both the old knowledge memorization and the new knowledge capturing through their coordination. Upon our proposed coordinated optimization scheme, we further enhance the anti-forgetting ability by adding KL consistency regularization $\mathcal{L}_{KL}$ (KL Reg.) and relational consistency regularization $\mathcal{L}_{Rel.}$. The comparison of the model \textit{Base.+CDR+RCL} to the model \textit{Base.+CDR} shows striking improvements of our proposed RCL, thanks to the loss function designs on old knowledge distillation/inheritance in line with the task objective of person ReID.

\begin{figure}[!t]
	\begin{center}
		\includegraphics[width=0.98\linewidth]{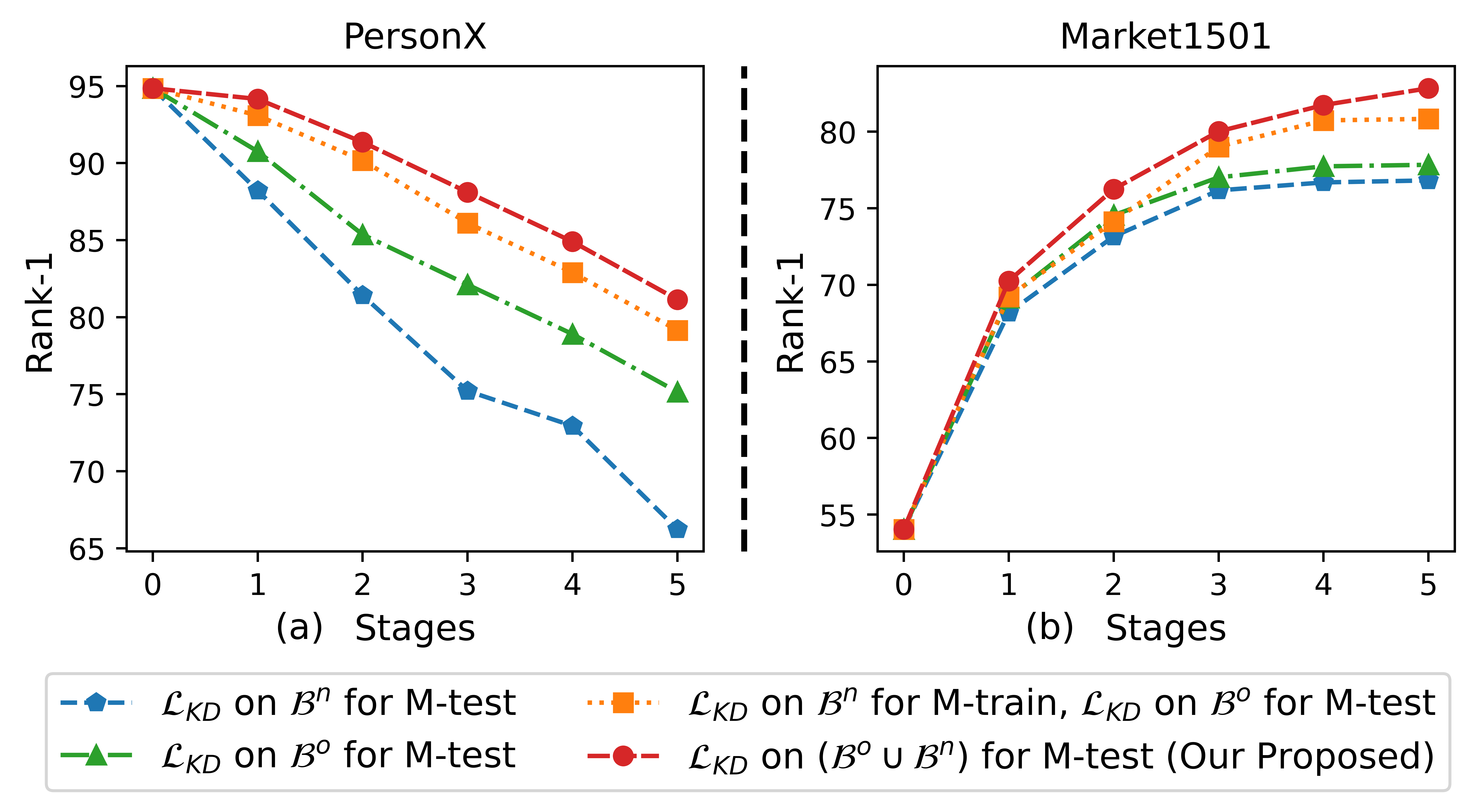}
	\end{center}
	\vspace{-0.50cm}
	\caption{The ablation study on the meta-optimization strategies. $\mathcal{L}_{KD}$ denotes the loss function for knowledge distillation. $\mathcal{B}^{o}$ and $\mathcal{B}^{n}$ denotes the batches of old data and new data, respectively. ``M-tarin'' is short for meta-train, ``M-test'' is short for meta-test. In all shown models, the person ReID losses ($\mathcal{L}_{Tri}$ and $\mathcal{L}_{Cls}$) applied on the new data are adopted for the meta-train while the person ReID losses applied to the old data are adopted for the meta-test. }
	\vspace{-0.50cm}
	\label{fig:ablation-meta}
\end{figure}

\noindent\textbf{Ablation study for meta-optimization strategies.} In this part, we empirically study the practices of using different data $\mathcal{B}^{o}$ and $\mathcal{B}^{n}$ for $\mathcal{L}_{KD}$ in meta-optimization. As shown in Fig.\ref{fig:ablation-meta}, by comparing our proposed scheme ($\mathcal{L}_{KD}$ on $\mathcal{B}^{o}\cup\mathcal{B}^{n}$ for meta-test) to the one applying $\mathcal{L}_{KD}$ to $\mathcal{B}^{o}$ for meta-test and the one applying $\mathcal{L}_{KD}$ to $\mathcal{B}^{n}$ for meta-test, we find that both $\mathcal{B}^{o}$ and $\mathcal{B}^{n}$ can contribute to the knowledge distillation for anti-forgetting. Besides, we further compare our proposed task-oriented meta-train/meta-test splitting in CDR with the data-based meta-train/meta-test splitting (\ieno, $\mathcal{L}_{KD}$ on $\mathcal{B}^{n}$ for meta-train while $\mathcal{L}_{KD}$ on $\mathcal{B}^{o}$ for meta-test) as in \cite{li2018learning,riemer2018learning}. We find our task-oriented splitting of meta-train and meta-test is more effective in coordinating the optimization objectives of anti-forgetting and adaptation.

\begin{figure}[!t]
	\begin{center}
		\includegraphics[width=0.98\linewidth]{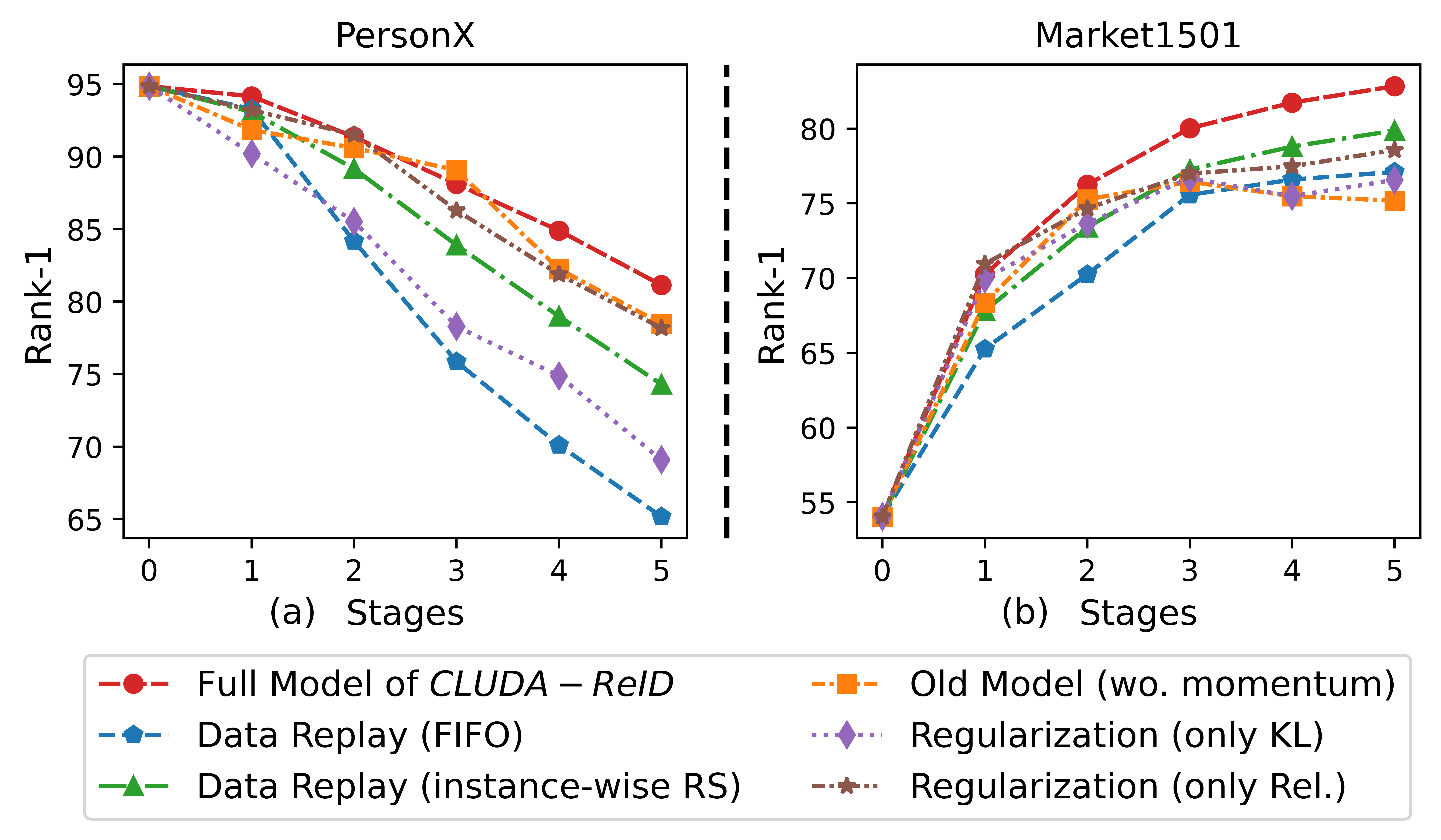}
	\end{center}
	\vspace{-0.50cm}
	\caption{The ablation study on different design choices for data replay, historical model updating and consistency regularization designs. This is conducted in the stationary target scenario. ``\textit{Data Replay (FIFO)}'' denotes the scheme storing seen samples into the memory buffer with the rule of First-In-First-Out. ``\textit{Instance-wise RS}'' refers to the regular version of Reservoir Sampling as introduced in \cite{vitter1985random}. ``\textit{Old Model (wo. momentum)}'' denotes we preserve a model at a certain old moment as the historical model. ``\textit{only KL}'' and ``\textit{only Rel.}'' means using $\mathcal{L}_{KL}$ and using $\mathcal{L}_{Rel.}$ as the knowledge distillation loss $\mathcal{L}_{KD}$, respectively.}
	\vspace{-0.50cm}
	\label{fig:design-choice}
\end{figure}

\noindent\textbf{Ablation study for different design choices.} We conduct an ablation study on different design choices in \ours~by replacing the components in the \textit{Full Model} of \ours~(the same as \textit{Base.+CDR+RCL} in Fig.\ref{fig:figure1}) with other design choices. The results are shown in Fig.\ref{fig:design-choice}. Compared to the First-In-First-Out (FIFO) rule and the regular (instance-wise) reservoir sampling algorithm, our modified ID-wise reservoir sampling adopted in \textit{Full Model} is superior since it enables stored samples to be diverse and ID-balanced. Compared to the historical model preserved at a certain historical moment, our design in \textit{Full Model} is more effective thanks to the temporal averaging via the momentum updating strategy. In terms of the loss function design for consistency learning, we find $\mathcal{L}_{KL}$ and $\mathcal{L}_{Rel.}$ are complementary for the old knowledge distillation from the historical model to the current model, since $\mathcal{L}_{KL}$ aims to learn absolute ID-related features while $\mathcal{L}_{Rel.}$ captures the relative information about inter-instance relations.

\begin{table*}[tb]
	\centering
	\footnotesize
	\vspace{-0.20cm}
	\resizebox{\linewidth}{!}{
		\begin{tabular}{l||cc||cc||cc||cc||cc||cc}
			\Xhline{3\arrayrulewidth}
			          \multicolumn{1}{c||}{\multirow{2}{*}{Methods}}& \multicolumn{2}{c||}{MA (t=1)} & \multicolumn{2}{c||}{SY (t=2)} & \multicolumn{2}{c||}{MS (t=3)} & \multicolumn{2}{c||}{MA (t=4)} & \multicolumn{2}{c||}{SY (t=5)} & \multicolumn{2}{c}{MS (t=6)} \\ \cline{2-13}
          & R-1 & mAP & R-1 & mAP & R-1 & mAP & R-1 & mAP & R-1 & mAP & R-1 & mAP \\ \Xhline{2\arrayrulewidth}
        Stage-wise UDA (Base.) & 72.23 & 49.32 & 64.59 & 60.76 & 19.80 & 7.24  & 72.92 & 50.46 & 63.93 & 60.29 & 20.23 & 7.18 \\
        Base. + Data Replay & 72.51 & 50.82 & 66.22 & 61.75 & 20.98 & 7.86  & 74.12 & 51.03 & 69.43 & 64.44 & 21.73 & 8.48 \\
        Base. + CDR & 73.12 & 51.11 & 69.23 & 64.54 & 23.56 & 9.33  & 76.93 & 53.46 & 73.46 & 69.29 & 25.90 & 10.43 \\
        Base. + CDR + KL Reg. & 74.24 & 51.76 & 71.55 & 67.88 & 26.16 & 9.94  & 79.26 & 59.48 & 76.76 & 74.45 & 29.79 & 12.44 \\
        \rowcolor[gray]{0.9}Base. + CDR + RCL  & \textbf{75.21} & \textbf{52.96} & \textbf{75.38} & \textbf{71.35} & \textbf{28.67} & \textbf{11.04} & \textbf{82.79} & \textbf{64.87} & \textbf{81.86} & \textbf{78.39} & \textbf{33.91} & \textbf{14.64} \\
        \hline
        All-in-one UDA & 85.12 & 68.59 & 86.76 & 84.70 & 36.37 & 17.84 & 85.12 & 68.59 & 86.76 & 84.70 & 36.37 & 17.84 \\
			\Xhline{3\arrayrulewidth}
		\end{tabular}
	}
	 \vspace{-0.20cm}
	 \caption{Adaptation performance (\%) evaluation in the dynamic target scenario. We test the model instantly at the end of each stage, on its corresponding test set. The training order is PX$\rightarrow$MA$\rightarrow$SY$\rightarrow$MS$\rightarrow$MA$\rightarrow$SY$\rightarrow$MS. Our \ours~is marked in gray shading.}
	 \vspace{-0.30cm}
	\label{table:adaptation}
\end{table*}

\begin{table*}[tb]
	\centering
	\footnotesize
	\resizebox{\linewidth}{!}{
		\begin{tabular}{l||cc|cc|cc||cc|cc|cc}
		\Xhline{3\arrayrulewidth}
          \multicolumn{1}{c||}{\multirow{3}{*}{Methods}} & \multicolumn{6}{c||}{t=3}                                       & \multicolumn{6}{c}{t=6} \\ \cline{2-13}
          & \multicolumn{2}{c|}{PX} & \multicolumn{2}{c|}{MA} & \multicolumn{2}{c||}{SY} & \multicolumn{2}{c|}{PX} & \multicolumn{2}{c|}{MA} & \multicolumn{2}{c}{SY} \\ \cline{2-13}
          & R-1    & mAP   & R-1    & mAP   & R-1    & mAP   & R-1    & mAP   & R-1    & mAP   & R-1    & mAP \\ \hline
    Stage-wise UDA (Base.) & 52.71 & 30.76 & 49.76 & 24.59 & 52.97 & 55.99 & 46.92 & 24.59 & 50.63 & 19.64 & 51.17 & 51.65 \\
    Base. + Data Replay & 56.33 & 38.21 & 57.46 & 31.73 & 54.46 & 56.33 & 48.78 & 27.33 & 60.34 & 32.37 & 58.93 & 57.11 \\
    Base. + CDR & 63.93 & 44.43 & 61.43 & 33.46 & 60.52 & 60.43 & 58.46 & 39.64 & 65.34 & 39.46 & 65.33 & 62.77 \\
    Base. + CDR + KL Reg. & 73.75 & 54.33 & 65.83 & 40.30 & 68.55 & 62.11 & 65.45 & 47.43 & 71.43 & 44.10 & 72.33 & 69.48 \\
    \rowcolor[gray]{0.9}Base. + CDR + RCL & \textbf{80.66} & \textbf{69.48} & \textbf{69.97} & \textbf{45.60} & \textbf{73.24} & \textbf{69.44} & \textbf{76.47} & \textbf{58.23} & \textbf{76.25} & \textbf{49.12} & \textbf{78.28} & \textbf{75.43} \\ \hline
    Pre-trained Model (t=0) & 94.86 & 86.34 & 54.04 & 27.67 & 46.52 & 48.62 & 94.86 & 86.34 & 54.04 & 27.67 & 46.52 & 48.62 \\
    All-in-one UDA & 40.43 & 16.44 & 85.12 & 68.59 & 86.76 & 84.70 & 40.43 & 16.44 & 85.12 & 68.59 & 86.76 & 84.70 \\

    \Xhline{3\arrayrulewidth}
    \end{tabular}%
	}
	\vspace{-0.20cm}
	\caption{Anti-forgetting performance (\%) evaluation in the dynamic target scenario. We test the model on the test-sets of all domains at the end of the 3rd stage (t=3) and the 6th stage (t=6). The training order is PX$\rightarrow$MA$\rightarrow$SY$\rightarrow$MS$\rightarrow$MA$\rightarrow$SY$\rightarrow$MS.  Note that we omit the MS dataset for brevity because it appears in the final, whose results can not reflect the anti-forgetting ability. \ours~is marked in gray.}
	\vspace{-0.40cm}
	\label{table:anti-forgetting}
\end{table*}

\subsection{Evaluation in the Dynamic Target Scenario}

\noindent\textbf{Experiment configuration.} We evaluate \ours~in the scenario with dynamic target stream introduced in Sec.\ref{subsec:setups}. This scenario is not only ID-incremental but also domain-incremental. One synthesis dataset PersonX (PX) is employed to simulate the source domain while three real-world datasets Market1501 (MA), CUHK-SYSU (SY), and MSMT17 (MS) are employed to simulate the dynamic target domain. We perform LUDA training on MA, SY, and MS domains in order, and repeat this process again to simulate the domain recurring cases in the real-world dynamic scenario. There are 6 stages in all for LUDA training in our experiment. Each stage includes a sub-set of 350 identities selected from the corresponding dataset via a random sampling without replacement. Furthermore, we also evaluate the model at the end of the final stage on MMP-Retrieval to assess its generalization ability on unseen domains. 

\noindent\textbf{Adaptation performance.} We evaluate the models at the end of each stage. As shown in Tab.\ref{table:adaptation}, compared to the scheme \textit{Stage-wise UDA (Base.)}, the scheme \textit{Base.+CDR} achieves 0.89\%/1.79\%, 4.64\%/3.78\%, 3.76\%/2.09\% improvements in R-1/mAP on MA, SY and MS respectively when these domains first appear, and achieves 4.01\%/3.00\%, 9.53\%/9.00\%, 5.67\%/3.25\% improvements in R-1/mAP on MA, SY and MS respectively when these domains show up for the second time. The scheme \textit{Base.+CDR+RCL} is superior to \textit{Stage-wise UDA} by 2.98\%/3.64\%, 10.79\%/10.59\%, 8.87\%/3.80\% respectively in R-1/mAP on MA, SY and MS when these domains first appear, by 9.87\%/14.41\%, 17.93\%/18.10\%, 13.68\%/7.46\% respectively in R-1/mAP on MA, SY and MS when they appear for the second time. These demonstrate the effectiveness of our proposed CDR and RCL in timely adapting to new environments. The improvements are especially significant when the similar or same domain re-appears thanks to the coordinated old knowledge memorization. 

\noindent\textbf{Anti-forgetting performance.} We measure the performance at the end of the 3rd stage ( \ieno, t=3, all domains are traversed once) and at the end of the 6th stage (\ieno, t=6, all domains are seen for the second time), for evaluating the anti-forgetting ability of our proposed method. As shown in Tab.\ref{table:anti-forgetting}, relative to the model pre-trained on PX, the scheme \textit{Base.+CDR+RCL} ranks the first with the lowest performance degradation on PX at the end of the 3rd and the 6th stages. Besides, comparing the results in Tab.\ref{table:anti-forgetting} to the timely measured results in Tab.\ref{table:adaptation}, we find the scheme \textit{Base.+CDR+RCL} is of the lowest performance degradation over all domains. It demonstrates the effectiveness of our \ours~in alleviating catastrophic forgetting. 

\begin{figure}[!t]
	\begin{center}
		\includegraphics[width=0.98\linewidth]{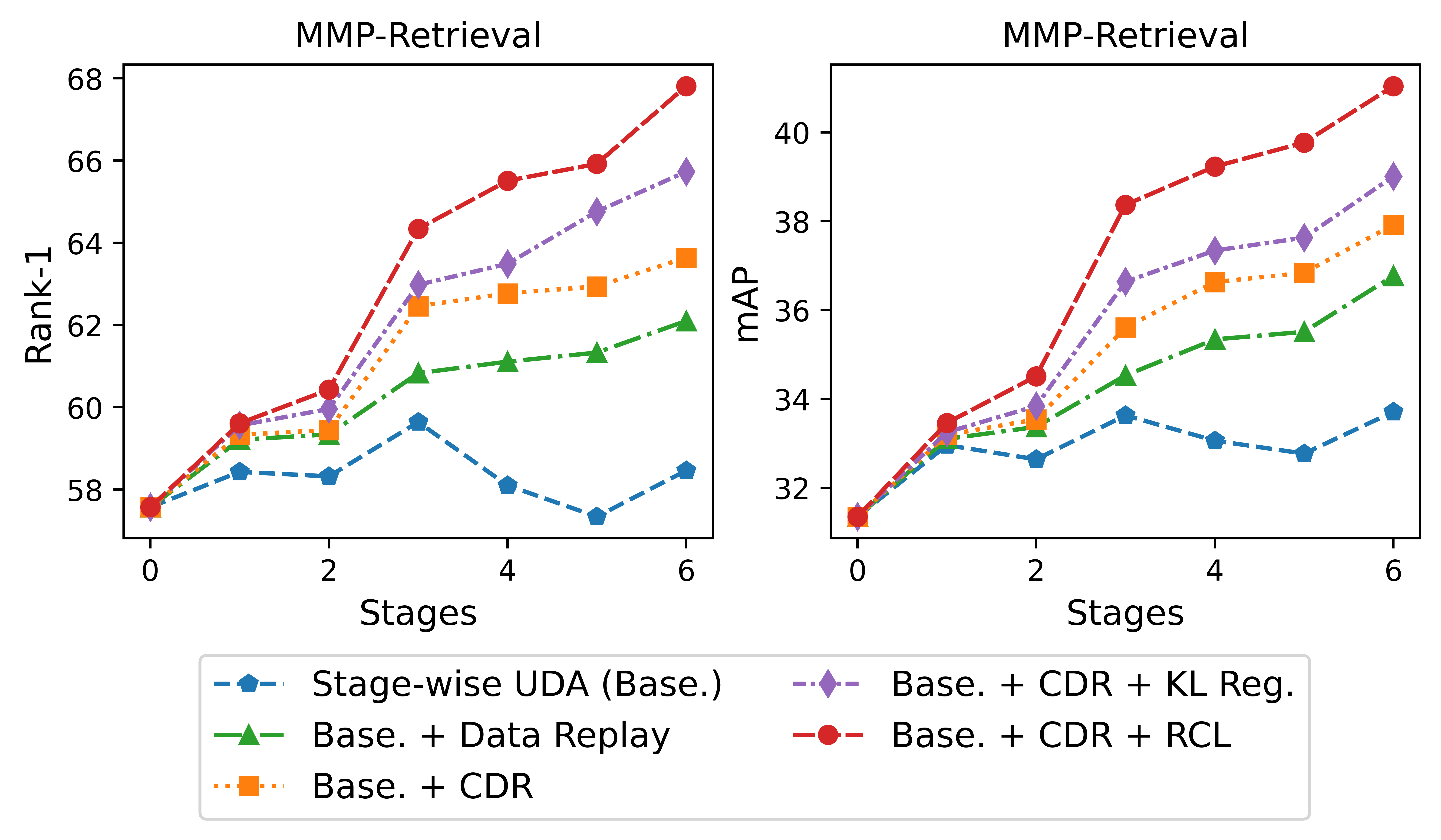}
	\end{center}
	\vspace{-0.50cm}
	\caption{Performance (\%) of unseen-domain generalization evaluation on our built MMP-Retrieval dataset.}
	\vspace{-0.50cm}
	\label{fig:unseen-domain}
\end{figure}

\noindent\textbf{Generalization performance on unseen domains.} We evaluate the trained model directly on MMP-Retrieval. As shown in Fig.\ref{fig:unseen-domain}, the Rank-1 and mAP of \textit{Base.+CDR} and \textit{Base.+CDR+RCL} on MMP-Retrieval increases constantly as the LUDA training goes on. This demonstrates the effectiveness of our proposed CDR and RCL of \ours~in improving the generalization ability for unseen domains through coordinated anti-forgetting and adaptation. 

\section{Conclusion}
In this paper, we propose a new and practical task, \ourtask, which aims to enable person ReID models to achieve continuous domain adaptation using unlabeled streaming data. This facilitates exploiting increased samples and achieves timely adaptation for real-world streaming data. Besides, it helps privacy protection of customer data as the practices in our experiments. For this task, we design an effective scheme \ours~in which the Coordinated Data Replay (CDR) and Relational Consistency Learning (RCL) are proposed to explicitly coordinate anti-forgetting and adaptation. We set up two practical evaluation schemes to simulate the real applications for method evaluation. Extensive experiment results demonstrate the effectiveness of \ours~in achieving LUDA for scenarios with stationary or dynamic target streams and enhancing the generalization capacity on unseen domains.

{\small
\bibliographystyle{ieee_fullname}
\bibliography{egbib}
}

\end{document}